# SOLUTION OF SYSTEM OF LINEAR EQUATIONS - A NEURO-FUZZY APPROACH

**Arindam Chaudhuri**[1]
**Lecturer & Research Fellow,**
**Mathematics and Computer Science,**
**Meghnad Saha Institute of Technology,**
**Nazirabad, Uchchepota, Kolkata, India**
**Email: arindam_chau@yahoo.co.in**
[1]**Corresponding Author**

**Kajal De**[2]
**Professor of Mathematics,**
**School of Science,**
**Netaji Subhas Open University,**
**Kolkata, India**

**Dipak Chatterjee**
**Distinguished Professor,**
**Department of Mathematics,**
**St. Xavier's College,**
**Kolkata, India**

## Abstract

Neuro-Fuzzy Modeling has been applied in a wide variety of fields such as Decision Making, Engineering and Management Sciences etc. In particular, applications of this Modeling technique in Decision Making by involving complex Systems of Linear Algebraic Equations have remarkable significance. In this Paper, we present Polak-Ribiere Conjugate Gradient based Neural Network with Fuzzy rules to solve System of Simultaneous Linear Algebraic Equations. This is achieved using Fuzzy Backpropagation Learning Rule. The implementation results show that the proposed Neuro-Fuzzy Network yields effective solutions for exactly determined, underdetermined and over-determined Systems of Linear Equations. This fact is demonstrated by the Computational Complexity analysis of the Neuro-Fuzzy Algorithm. The proposed Algorithm is simulated effectively using MATLAB software. To the best of our knowledge this is the first work of the Systems of Linear Algebraic Equations using Neuro-Fuzzy Modeling.

***Keywords*:** System of Linear Algebraic Equations, Neuro-Fuzzy Networks, Fuzzy Backpropagation Learning Rule, Polak-Ribiere Conjugate Gradient Algorithm

## 1. INTRODUCTION

Several problems in Engineering Sciences involve solving Systems of Linear Algebraic Equations, specially, in Signal Processing and Robotics. In Principle, solving Systems of Linear Equations is equivalent to computing the Inverse of a Matrix [3], [4]. The Numerical Methods provides excellent Algorithms for solving Systems of Linear Equations. Typically, the time constraints for solving Systems of Linear Equations in Non Real Time Systems are not important. However, in Real Time Systems the time constraints for solving these Systems of Linear Equations are more demanding than a Digital Machine can provide and it becomes necessary to find an alternate method of this solution. One approach is to use Neuro-Fuzzy Networks [8] that can learn as well as allow prior knowledge to be embedded via Fuzzy Rules [7] with appropriate linguistic labels because of their inherent parallel architecture [5].

[2]**Arindam Chaudhuri is working towards his PhD under the supervision of Dr. Kajal De**

The first phase consists of constructing an appropriate Error Cost Function for the particular type of problem to be solved. The Error Cost Function is based on defined error variables that are typically formulated from a Functional Network for the particular problem. Thus, the problem in general is represented by a structured Multilayer Neural Network [1], [2], [5], [9]. The Multilayer Neural Network has Fuzzy rules embedded in it giving rise to a Neuro-Fuzzy Network. The node and links in a Neuro-Fuzzy Network correspond to a specific component in a Fuzzy System. A node in a Neuro-Fuzzy Network is usually not fully connected to nodes in an adjacent layer. The second phase is an optimization step, which involves deriving the appropriate Learning Rule [5] for the Structured Neuro-Fuzzy Network using the defined Error cost function. This typically involves by deriving the Learning Rule in its Vector-Matrix form [6]. Once the Vector-Matrix form of Learning Rule is derived, the scalar form can be formulated in a relatively straightforward manner. The third phase involves the training of the Neuro-Fuzzy Network using the Learning Rule developed as above to match some set of desired Patterns i.e., Input/Output signal pairs [8]. Therefore, the Network is essentially optimized to minimize the associated Error Cost Function. That is, the Training phase involves adjusting the Network's synaptic weights [2] according to the derived Learning Rule in order to minimize the associated Error Cost Function. The nodes in different layers of a Neuro-Fuzzy Network perform different operations. The fourth and final phase is actually the Application phase in which appropriate output signals are collected from the structural Neural Network for a particular set of inputs [5] to solve a specific problem.

The rest of the paper is organized as follows: In Section 2, the Systems of Linear Algebraic Equations is presented. The Basic Concepts, Algorithms and Architecture of the Neuro-Fuzzy Network for solving the Systems of Linear Algebraic Equations follow this. In the next Section some Numerical Examples for exactly determined, underdetermined and over-determined Systems of Linear Equations are presented. This is followed by Complexity Analysis of the proposed Neuro-Fuzzy Algorithm. In Section 6, Conclusion is given.

## 2. SYSTEMS OF SIMULTANEOUS LINEAR ALGEBRAIC EQUATIONS

Given the set of linear algebraic equations with constant coefficients [10], [12],

$$
\begin{aligned}
a_{11}x_1 + \ldots\ldots\ldots\ldots + a_{1n}x_n &= b_1 \\
a_{21}x_1 + \ldots\ldots\ldots\ldots + a_{2n}x_n &= b_2 \\
&\ldots\ldots\ldots\ldots\ldots\ldots \\
&\ldots\ldots\ldots\ldots\ldots\ldots \\
a_{m1}x_1 + \ldots\ldots\ldots\ldots + a_{mn}x_n &= b_m
\end{aligned}
\quad (1)
$$

It is desired to find the unknown vector $X = [x_1 \ldots\ldots\ldots x_n]$ given the coefficients $a_{ij}; i = 1,\ldots\ldots,m, j = 1,\ldots\ldots,n$ and $b_i; i = 1,\ldots\ldots,m$. The System of Linear Equations (1) can be written in a more compact vector-matrix form as,

$$AX = B \quad (2)$$

where, it is assumed that $A \in R^{m \times n}, X \in R^{n \times 1}, B \in R^{m \times 1}$ and matrix $A$ and $B$ are given by:

$$A = \begin{bmatrix} 1 & . & . & . & 1 \\ . & . & . & . & . \\ . & . & . & . & . \\ . & . & . & . & . \\ 1 & . & . & . & 1 \end{bmatrix}_{m \times n} \qquad B = \begin{bmatrix} b_1 & . & . & . & b_m \end{bmatrix}^T$$

There are three cases that can exist viz., (i) If $n = m$ i.e. there are as many equations as unknowns. Then the system of equations are called exactly determined. (ii) If $n > m$ i.e. there are more unknowns than equations. Then the system of equations are called underdetermined. (iii) If $m > n$ i.e. there are more equations than unknowns. Then the system of equations are called overdetermined. This is a common situation encountered often in practice. An efficient way to solve these System of Linear Equations numerically is given by Gauss Jordan Elimination or by Cholesky Decomposition [4], [12]. For problems of the form in Equation (2), where $A$ is a Singular Matrix or nearly Singular; matrix $A$ is decomposed into product of three matrices in a process called Singular Value Decomposition. The types of systems that are of major interest here is to solve System of Equations that are much more complex, large-scale overdetermined and underdetermined systems, ill-conditioned systems and systems that have uncertainty associated with them. These complex systems are often tackled using Neuro-Fuzzy Network techniques [8].

## 3. BASIC CONCEPTS, ALGORITHMS AND ARCHITECTURE OF NEURO-FUZZY NETWORK

Neural Networks [2], [5] are computational models that consist of nodes that are connected by links. Each node performs a simple operation to compute its output from its input, which is transmitted through links connected to other nodes. This relatively simple computational model is Artificial Neural Network because the structure is analogous to that of Neural Systems in Human Brains – nodes corresponding to neurons and links corresponding to synapses that transmit signal between neurons. One of the major features of a Neural Network is its learning capability. While the details of learning algorithms of Neural Networks vary from architecture to architecture, they have one thing in common; they can adjust the parameters in a Neural Network such that the network learns to improve its performance of a given task.

Fuzzy Logic [11], [14] is a computational paradigm that generalizes classical two-valued logic for reasoning under uncertainty. In order to achieve this, the notation of membership in a set needs to become a matter of degree. This is the essence of Fuzzy Sets. By doing this one accomplishes two things (i) ease of describing human knowledge involving vague concepts and (ii) enhanced ability to develop a cost-effective solution to real-world problem. Fuzzy Logic is a kind of multi-valued logic, which is a model-less approach and is a clever disguise of the Probability Theory.

Neural Network and Fuzzy Logic are two complementary technologies. Neural Networks can learn from data and feedback; however, understanding the knowledge or the pattern learned by the Neural Network has been difficult. More specifically, it is difficult to develop an insight about the meaning associated with each neuron and each weight. Hence, Neural Networks are often viewed as a Black Box approach. In contrast, Fuzzy Rule-Based Models are easy to comprehend because it uses linguistic terms and the structure of if-then rules. Unlike Neural Network, Fuzzy Logic does not come with a learning algorithm. The learning and identification of Fuzzy Models adopts techniques from other areas such as Statistics, Linear System Identification etc. Since, Neural Networks can learn, it is natural to merge the two technologies. This merger has created a new term Neuro-Fuzzy Networks [7], [8]. A Neuro-Fuzzy Network thus describes a Fuzzy Rule-Based Model using a Neural Network like structure.

**3.1 Neuro-Fuzzy Network Approach**

We find a solution for Equation (2) that is based on Neuro-Fuzzy Network Approach [7], [8] that can learn as well as allow prior knowledge to be embedded via Fuzzy Rules with appropriate linguistic labels. Typically, a Neuro-Fuzzy Network has five to six layers of nodes. The functionalities associated with different layers include the following:

1. Compute the matching degree to a Fuzzy condition involving one variable.
2. Compute the matching degree to a conjunctive Fuzzy condition involving multiple variables.
3. Compute the normalized matching degree.
4. Compute the conclusion inferred by a Fuzzy rule.
5. Combine the conclusion of all Fuzzy rules in a model.

Here, our approach is based on Fuzzy Backpropagation Learning Rule.

**3.2 Fuzzy Back-propagation Learning Rule**

The Fuzzy Backpropagation Learning Rule i.e. Backpropagation Learning [5], [13] applied to Fuzzy Modeling is developed to solve the Equation (2). The Backpropagation Learning is generally obtained by applying the Conjugate Gradient Descent Method to minimize the error between the network's output and target output of the entire training set. In Conjugate Gradient Descent Method, a set of direction vectors $\{d_0, \ldots\ldots\ldots, d_{n-1}\}$ is generated that are conjugate with respect to the matrix $A$ i.e. $d_i^T A d_j = 0; i \neq j$.

A conjugate direction vector is generated at $k^{th}$ iteration of the iterative process by adding to the calculated current negative gradient vector of the Objective Function. The conjugate direction vector generated at $k^{th}$ iteration is applied to the parameter identification of TSK (Takagi-Sugeno-Kang) Fuzzy Model [2] whose antecedent membership functions are of Gaussian type. More specifically, in the forward pass for a given input pattern, the actual response of the model is computed directly from Equation (6), and the effect from the input to the output is completed through just a single propagation step. During this process, the antecedent parameters $m_{ij}$, $\sigma_{ij}$ and consequents $c_i$, which amount to the weights in the Neural Network, are all fixed. In the backward pass, the Error Signal resulting from the difference between the actual output and the desired output is propagated backward and the parameters $m_{ij}$, $\sigma_{ij}$ and $c_i$ are adjusted using Error Correction Rule. Again, the process is completed in a single propagation step. Denoting the Error Function at $k^{th}$ iteration as [5], [6],

$$J(k) = \frac{1}{2}(\hat{y}_i - y_i)^2 \quad (3)$$

The Error Correction rules for $c_i$, $m_{ij}$ and $\sigma_{ij}$ are given by,

$$c_i(k) = c_i(k-1) - \eta_1 \left(\frac{\partial J(k)}{\partial c_i}\right)|_{c_i = c_i(k-1)}; i = 1, \ldots\ldots, M \quad (4)$$

$$m_{ij}(k) = m_{ij}(k-1) - \eta_2 (\frac{\partial J(k)}{\partial m_{ij}})|_{m_{ij}=m_{ij}(k-1)}; i = 1,\ldots\ldots,M, j = 1,\ldots\ldots,m \quad (5)$$

$$\sigma_{ij}(k) = \sigma_{ij}(k-1) - \eta_3 (\frac{\partial J(k)}{\partial \sigma_{ij}})|_{\sigma_{ij}=\sigma_{ij}(k-1)}; i = 1,\ldots\ldots,M, j = 1,\ldots\ldots,m \quad (6)$$

where, $\eta_1, \eta_2, \eta_3$ are learning rate parameters. To give a clear picture that shows how gradients $(\frac{\partial J(k)}{\partial c_i})$, $(\frac{\partial J(k)}{\partial m_{ij}})$ and $(\frac{\partial J(k)}{\partial \sigma_{ij}})$ are formed and $\hat{y}(k)$, actual output of model at $k^{th}$ observation in its extended are rewritten as:

$$\hat{y}(\hat{k}) = \Phi(\hat{k})\theta = \sum_{i=i}^{M} v_i(k)c_i \quad (7)$$

$$\text{where, } v_i(k) = \frac{\prod_{j=1}^{S} \mu_{A_{ij}} x_j(k)}{\sum_{i=1}^{M}\prod_{j=1}^{S} \mu_{A_{ij}} x_j(k)} = \frac{\prod_{j=1}^{S} \exp((x_j(k)-m_{ij})^2/\sigma_{ij}^2)}{\sum_{i=1}^{M}\prod_{j=1}^{S}((x_j(k)-m_{ij})^2/\sigma_{ij}^2)} \quad (8)$$

Correspondingly error signal becomes,

$$e(k) = y(k) - \hat{y}(k) = y(k) - \sum_{i=1}^{M} v_i(k)c_i \quad (9)$$

Now, differentiating $J(k)$ with respect to $\hat{\theta}$, equating result to zero and using chain rule,

$$\frac{\partial J(k)}{\partial c_i} = \frac{\partial J(k)\partial e(k)}{\partial e(k)\partial c_i} = -2e(k)v_i(k) \quad (10)$$

$$\frac{\partial J(k)}{\partial m_{ij}} = \frac{\partial J(k)\partial e(k)\partial v_i(k)}{\partial e(k)\partial v_i(k)\partial m_{ij}} = -2e(k)v_i(k)[c_i - \sum_{l=1}^{M} v_l(k)c_l][((x_j(k)-m_{ij})/\sigma_{ij}^2] \quad (11)$$

$$\frac{\partial J(k)}{\partial \sigma_{ij}} = \frac{\partial J(k)\partial e(k)\partial v_i(k)}{\partial e(k)\partial v_i(k)\partial \sigma_{ij}} = -2e(k)v_i(k)[c_i - \sum_{l=1}^{M} v_l(k)c_l][((x_j(k)-m_{ij})^2/\sigma_{ij}^3] \quad (12)$$

Substituting Equations (10), (11) and (12) into Equations (13), (14) and (15) respectively,

$$c_i(k) = c_i(k-1) + 2\eta_1 e(k)v_i(k); i = 1,\ldots\ldots,M \quad (13)$$

$$m_{ij}(k) = m_{ij}(k-1) + 2\eta_2 e(k)v_i(k)[c_i - \sum_{l=1}^{M} v_l(k)c_l(k)][((x_j(k)-m_{ij}(k))/\sigma_{ij}^2(k))]; \quad (14)$$
$$i = 1,\ldots\ldots,M, j = 1,\ldots\ldots,S$$

$$\sigma_{ij}(k) = \sigma_{ij}(k-1) + 2\eta_3 e(k)v_i(k)[c_i - \sum_{l=1}^{M} v_l(k)c_l(k)][((x_j(k)-m_{ij}(k))^2/\sigma_{ij}^3(k))]; \quad (15)$$
$$i = 1,\ldots\ldots,M, j = 1,\ldots\ldots,S$$

where, $v_i(k)$ are computed by Equation (17) with $m_{ij}$ and $\sigma_{ij}$ replaced by $m_{ij}(k-1)$ and $\sigma_{ij}(k-1)$ respectively and $e(k)$ is computed by Equation (18) with $c_i$ replaced by $c_i(k-1)$. The last three equations make up the back-propagation algorithm for parameter estimation of TSK fuzzy models using gaussian antecedent membership functions. This is valid for purely quadratic case. However, it is important for generalizations of conjugate gradient method to non-quadratic problems. It is assumed that near solution point the problem is approximately quadratic. Here, polak-ribiere conjugate gradient method [2], [9] based on line search methods for non-quadratic case is presented. The method is modified using fuzzy rules with appropriate linguistic labels. The objective is to minimize cost function $E(x)$ where $x \in R^{n \times 1}$ and $E$ is not necessarily quadratic function.

**3.3 Polak-Ribiere Conjugate Gradient Algorithm with Fuzzy Rules**

**Step 1:** Set $x_0$

**Step 2:** Compute $g_0 = \nabla_x E(x_0) = \partial E(x)/\partial x$ at $x = x_0$

**Step 3:** Set $d_0 = -g_0$

**Step 4:** Compute $x_{k+1} = x_k + \alpha_k d_k v_k$ where, $x_{k+1} = \min_{\alpha \geq 0} E(x_k + \alpha_k d_k v_k)$

and $v_k = \dfrac{\prod_{j=1}^{S} \mu_{A_{ij}} x_j(k)}{\sum_{i=1}^{M} \prod_{j=1}^{S} \mu_{A_{ij}} x_j(k)}$

where, $\beta_k = g_{k+1}^T (g_{k+1} - g_k)/g_k^T g_k$

**Step 5:** Compute $g_{k+1} = \nabla_x E(x_{k+1})$

**Step 6:** Compute $d_{k+1} = -g_{k+1} + \beta_k d_k$

Step 4 through Step 6 are carried out for $k = 0,1,\ldots\ldots\ldots,n-1$

**Step 7:** Replace $x_0$ by $x_n$ and go to Step1

**Step 8:** Continue until convergence is achieved; termination criterion could be $\| d_k \| < \varepsilon$

(where, $\varepsilon$ is an appropriate predefined small number)

The restart feature in the above Algorithm (Step 7) is important for the cases where the Cost Function is not quadratic. The Polak-Ribiere Conjugate Gradient Algorithm with Fuzzy Rules is restated by a search in the steepest descent directions after each $n$ iteration [2], [11]. The restart feature of the Algorithm is important for global convergence because one cannot guarantee that the directions that $d_k$ generate are descent directions.

**3.4 Architecture of the Neuro-Fuzzy Network**

Modified polak-ribiere conjugate gradient algorithm with fuzzy rules [8], [11] gives the updated solution,

$$x_{k+1} = x_k + \alpha_k d_k v_k \text{ where } x_{k+1} = \min_{\alpha \geq 0} E(x_k + \alpha_k d_k v_k)$$

The vector $d_k$ is current direction vector and $E(\bullet)$ is cost function to be minimized. In this case cost function is given by,

$$E(x) = \tfrac{1}{2} \| Ax - b \|^2 = \tfrac{1}{2}(Ax-b)^T (Ax-b) \qquad (16)$$

Therefore, $E(x_k + \alpha d_k v_k) = \frac{1}{2}[A(x_k + \alpha d_k v_k) - b]^T [A(x_k + \alpha d_k v_k) - b]$

$E(x_k + \alpha d_k v_k) = \frac{1}{2}[(x_k + \alpha d_k v_k)^T A^T - b^T][Ax_k + \alpha A d_k v_k - b]$

$E(x_k + \alpha d_k v_k) = \frac{1}{2}[(x_k^T + \alpha d_k^T v_k^T)A^T - b^T][Ax_k + \alpha A d_k v_k - b]$

$E(x_k + \alpha d_k v_k) = \frac{1}{2}[(x_k^T A^T + \alpha d_k^T v_k^T A^T - b^T][Ax_k + \alpha A d_k v_k - b]$

$E(x_k + \alpha d_k v_k) = \frac{1}{2}[x_k^T A^T A x_k + \alpha x_k^T A^T A d_k v_k - x_k^T A^T b + \alpha d_k^T v_k^T A^T A x_k$

$+ \alpha^2 d_k^T v_k^T A^T A d_k v_k - \alpha d_k^T v_k^T A^T b - b^T A x_k - \alpha b^T A d_k v_k - b^T b]$ (17)

Computing the gradient of $E(x_k + \alpha d_k v_k)$ in Equation (17) with respect to $\alpha$ and equating the result equal to zero,

$$\nabla_\alpha E(x_k + \alpha d_k v_k) = \partial E(x_k + \alpha d_k v_k)/\partial \alpha$$

$\nabla_\alpha E(x_k + \alpha d_k v_k) = \frac{1}{2}[x_k^T A^T A d_k v_k + d_k^T v_k^T A^T A x_k + 2\alpha d_k^T v_k^T A^T A d_k v_k - d_k^T v_k^T A^T b - b^T A d_k v_k]$

$\nabla_\alpha E(x_k + \alpha d_k v_k) = \frac{1}{2}[2 d_k^T A^T A x_k v_k + 2\alpha d_k^T v_k^T A^T A d_k v_k - 2 d_k^T A^T b v_k]$

$\nabla_\alpha E(x_k + \alpha d_k v_k) = 0$

$d_k^T A^T A x_k v_k + \alpha d_k^T v_k^T A^T A d_k v_k - d_k^T A^T b v_k = 0$ (18)

Solving for $\alpha$ from Equation (18) and assuming $\alpha = \alpha_k$ gives,

$$\alpha_k = -\frac{g_k^T v_k^T d_k}{d_k^T v_k^T A^T A d_k}$$ (19)

Here, $g_k v_k$ is gradient of $E(x_k)$ i.e., $g_k v_k = \nabla_x E(x_k)$ or $g_k = \nabla_x E(x_k)/v_k$ (20)

Thus, modified polak-ribiere conjugate gradient algorithm with fuzzy rules (with restart) [8], [11] for solving $AX = B$ is summarized in following steps:

**Algorithm**

**Step 1:** Set $x_0$

**Step 2:** Compute $g_0 = (A^T A x_0 - A^T b)/v_0$

**Step 3:** Set $d_0 = -g_0$

**Step 4:** Compute $x_{k+1} = x_k + \alpha_k d_k v_k$ where, $\alpha_k = -\dfrac{g_k^T v_k^T d_k}{d_k^T v_k^T A^T A d_k}$

**Step 5:** Compute $g_{k+1} = (A^T A x_{k+1} - A^T b)/v_{k+1}$

**Step 6:** Compute $d_{k+1} = -g_{k+1} + \beta_k d_k$ where $\beta_k = g_{k+1}^T(g_{k+1} - g_k)/g_k^T g_k$

Step 4 through Step 6 are carried out for $k = 0, 1, \ldots\ldots\ldots, n-1$

**Step 7:** Replace $x_0$ by $x_n$ and go to Step 1

**Step 8:** Continue until convergence is achieved; termination criterion could be $\| d_k \| < \varepsilon$
(where, $\varepsilon$ is an appropriate predefined small number)

## 4. NUMERICAL EXAMPLES

In order to verify the effectiveness of Neuro–Fuzzy Network approach, some simple examples are simulated using MATLAB software to solve exactly determined, underdetermined and over–determined System of Linear Algebraic Equations [3], [4], [10], [12].

*Example* 1

Consider the following exactly determined system of linear equations:

$$10x_1 - 2x_2 - x_3 - x_4 = 3$$
$$-2x_1 + 10x_2 - x_3 - x_4 = 15$$
$$-x_1 - x_2 + 10x_3 - 2x_4 = 27$$
$$-x_1 - x_2 - 2x_3 + 10x_4 = -9$$

The neuro-fuzzy conjugate gradient algorithm given above is used to solve this system of equations for $X = [x_1, x_2, x_3, x_4]$. The initial zero condition is assumed for unknown quantities. After $3^{rd}$ iteration, the solution is given as $X = [1,2,3,0]$. The same solution is obtained using gauss–seidel method after 7 iterations and jacobi method after 12 iterations.

*Example* 2

Consider the following exactly determined system of linear equations:

$$20x_1 + x_2 - 2x_3 = 17$$
$$3x_1 + 20x_2 - x_3 = -18$$
$$2x_1 - 3x_2 + 20x_3 = 25$$

The neuro-fuzzy conjugate gradient algorithm given above is used to solve this system of equations for $X = [x_1, x_2, x_3]$. The initial zero initial condition is assumed for unknown quantities. After $4^{th}$ iteration, the solution is given as $X = [1,-1,1]$. The same solution is obtained using jacobi method after 6 iterations.

*Example* 3

Consider the following underdetermined system of linear equations:

$$6x_1 + 2x_2 + 4x_3 - 9x_4 - 12x_5 + 2x_6 - 12x_7 + x_9 = -12$$
$$8x_1 - 10x_2 + x_3 + 8x_4 - 22x_5 - 11x_7 - 11x_8 + 7x_9 = -13$$
$$9x_1 - 7x_2 - 6x_3 + 6x_4 + 10x_5 - 10x_6 + 15x_7 - 13x_8 - 12x_9 = 9$$
$$-10x_1 + 11x_2 - 6x_3 - 8x_4 - 5x_5 - 9x_6 + x_7 - 3x_8 - 5x_9 = 0$$
$$2x_1 - x_2 + 4x_3 - 3x_4 + 3x_5 - 4x_6 - 12x_7 + 10x_8 - 3x_9 = -6$$

The neuro-fuzzy conjugate gradient algorithm given above is used to solve this system of equations for $X = [x_1, x_2, x_3, x_4, x_5, x_6, x_7, x_8, x_9]$. The initial zero condition is assumed for unknown quantities. The proposed algorithm gives the following solution after $4^{th}$ iteration:
$$X = [-0.1886, 0.4444, -0.1066, 0.1450, 0.3418, -0.0678, 0.4396, -0.0186, 0.0060]$$
The same solution is obtained using singular-value decomposition method.

*Example* 4

Consider the following over-determined system of linear equations:

$$x_1 + 2x_2 + 3x_3 = 14$$
$$3x_1 + 2x_2 + x_3 = 10$$
$$x_1 + x_2 + x_3 = 6$$
$$2x_1 + 3x_2 - x_3 = 5$$
$$x_1 + x_2 = 3$$

The neuro-fuzzy conjugate gradient algorithm solves this system of equations for $X = [x_1, x_2, x_3]$. The zero initial condition is assumed for unknown quantities. The proposed algorithm gives solution $X = [1, 2, 3]$ after $2^{nd}$ iteration.

## 5. COMPLEXITY ANALYSIS OF THE MODIFIED POLAK-RIBIRE CONJUGATE GRADIENT NEURO-FUZZY ALGORITHM

In the context of analyzing the Computational Complexity of the Modified Polak-Ribiere Conjugate Gradient Algorithm with Fuzzy Rules for solving the System of Simultaneous Linear Algebraic Equations we express the Linear System as a Matrix Equation $AX = B$ [3], [4], [10], [12] in which each Matrix belongs to a Field of Real Numbers $\Re$. Further we consider three cases when the System of Linear Equations is exactly determined, underdetermined and over-determined. Now the central question is "How fast the Solution of Simultaneous Linear Algebraic Equations be solved using Modified Polak-Ribiere Conjugate Gradient Algorithm with Fuzzy Rules?" or "What is the Computational Complexity of the above Algorithm?"

For the exactly determined case in which the number of equations is equal to the number of unknowns, using the above Neuro-Fuzzy Algorithm the Computational Complexity is $\Theta(n^2)$ which defines the asymptotic tight bound. Again, considering the underdetermined case in which the number of equations is less than the number of unknowns, typically such a system has infinitely many solutions (if there are any). However, using the above Neuro-Fuzzy Algorithm the Computational Complexity is $O(n^2)$ which defines the asymptotic upper bound. Finally, considering the over-determined case in which the number of equations exceeds the number of unknowns, there may not exist any solution. Finding good approximate solutions to over-determined Systems of Linear Equations is an important problem. Here, the above Neuro-Fuzzy Algorithm yields the Computational Complexity $\Omega(n^2)$ which defines the asymptotic lower bound. Thus, from the above Computational Complexity discussion we conclude that the proposed Neuro-Fuzzy Algorithm provides improved results over other well-known techniques.

## 6. CONCLUSION

The major objective of this work is to present the concepts of Neural Networks and Fuzzy Logic to develop the Neuro-Fuzzy Algorithm using Polak-Ribiere Conjugate Gradient Method to solve the Systems of Linear Equations for all the three cases, namely exactly determined, underdetermined and over-determined. This is achieved using Fuzzy Backpropagation Learning Rule. The Neuro-Fuzzy Algorithm presented in this paper has several advantages over the conventional Neural Network Algorithms. First, the Neuro-Fuzzy Algorithms are much simpler than others. The only algebraic operations required here are addition and multiplication. Inverse and other complex Logic operations are not needed. Secondly, the Neuro-Fuzzy Algorithms include the most parallel and distributed attribute of the conventional Neural Network Algorithms. Thirdly, the nodes and links in a Neuro-Fuzzy Network are comprehensible. Fourthly, a Neuro-Fuzzy Network has more layers than the conventional Neural Networks which allow greater degree of precision in solution obtained. Finally, the Neuro-Fuzzy Algorithms are simple to implement. We have presented various Numerical Examples to prove the effectiveness of the Neuro-Fuzzy Algorithm developed. A discussion on the Computational Complexity of the proposed Neuro-Fuzzy Algorithm is also given.

**Acknowledgements**

We wish to thank all Technical Staff of Techno India Group, Kolkata and St. Xavier's College, Kolkata for their constant support and encouragement in presenting this paper.